\documentclass[letterpaper, 10 pt, conference]{ieeeconf}  

\IEEEoverridecommandlockouts                              

\usepackage{amssymb,graphicx,amsmath,color,algorithm,algorithmic,lipsum,subcaption,tabularx,soul}
\usepackage[noadjust]{cite} 
\title{Voronoi Features for Tactile Sensing:\\ Direct Inference of Pressure, Shear, and Contact Locations  
\vspace{0em}
}
\author{Luke Cramphorn, John Lloyd, Nathan F. Lepora%
\thanks{The work of N.~Lepora was supported in part by Leadership Award from the Leverhulme Trust on `A biomimetic forebrain for robot touch' (RL-2016-39). The work of L.~Cramphorn was supported by a studentship in the EPSRC Centre for Doctoral Training in Future Autonomous and Robotic Systems (FARSCOPE)}
\thanks{The authors are with Bristol Robotics Laboratory, and the Department of Engineering Mathematics, University of Bristol, Bristol, U.K. \newline
	Email: \{ll14468, jl15313, n.lepora\}@bristol.ac.uk}
}
	
\begin{document}

\maketitle

\begin{abstract}  
There are a wide range of features that tactile contact provides, each with different aspects of information that can be used for object grasping, manipulation, and perception. In this paper inference of some key tactile features, tip displacement, contact location, shear direction and magnitude, is demonstrated by introducing a novel method of transducing a third dimension to the sensor data via Voronoi tessellation. The inferred features are displayed throughout the work in a new visualisation mode derived from the Voronoi tessellation; these visualisations create easier interpretation of data from an optical tactile sensor that measures local shear from displacement of internal pins (the TacTip). The output values of tip displacement and shear magnitude are calibrated to appropriate mechanical units and validate the direction of shear inferred from the sensor. We show that these methods can infer the direction of shear to $\sim$2.3$^{\circ}$ without the need for training a classifier or regressor. The approach demonstrated here will increase the versatility and generality of the sensors and thus allow sensor to be used in more unstructured and unknown environments, as well as improve the use of these tactile sensors in more complex systems such as robot hands.     
\end{abstract}


\section{INTRODUCTION}

\PARstart{T}{actile} perception is an important ability for any manipulator, whether that be a human or robotic hand. The ability to interpret key tactile features is crucial for making decisions on the perception and manipulation of objects and environments. An increasingly popular tool for these sensory task in robotics are optical tactile sensors. The reason for this being that such sensors like the TacTip, GelSight~\cite{Li2014} and MIS sensor from KCL~\cite{Back2015} are low cost and customizable, as well as robust with high acuity.

In optical tactile sensors, a common method to transduce deformation of the sensing surface is to fix markers and track them. However, the tracked position of these markers are not directly linked with the physical dimensions of touch. This necessitates the use of black box classification or regression techniques, such as neural networks or Bayesian methods, when utilising the system with robots. While able to perform complex and highly accurate open and closed loop tasks, these methods suffer from being non-transparent in how they work, time consuming to train, and limit the ability of the system to generalise for other tasks.

\begin{figure}[t!]
	\centering
	{\includegraphics[width=0.42\textwidth, trim = 0cm 1cm 0cm 1cm, clip]{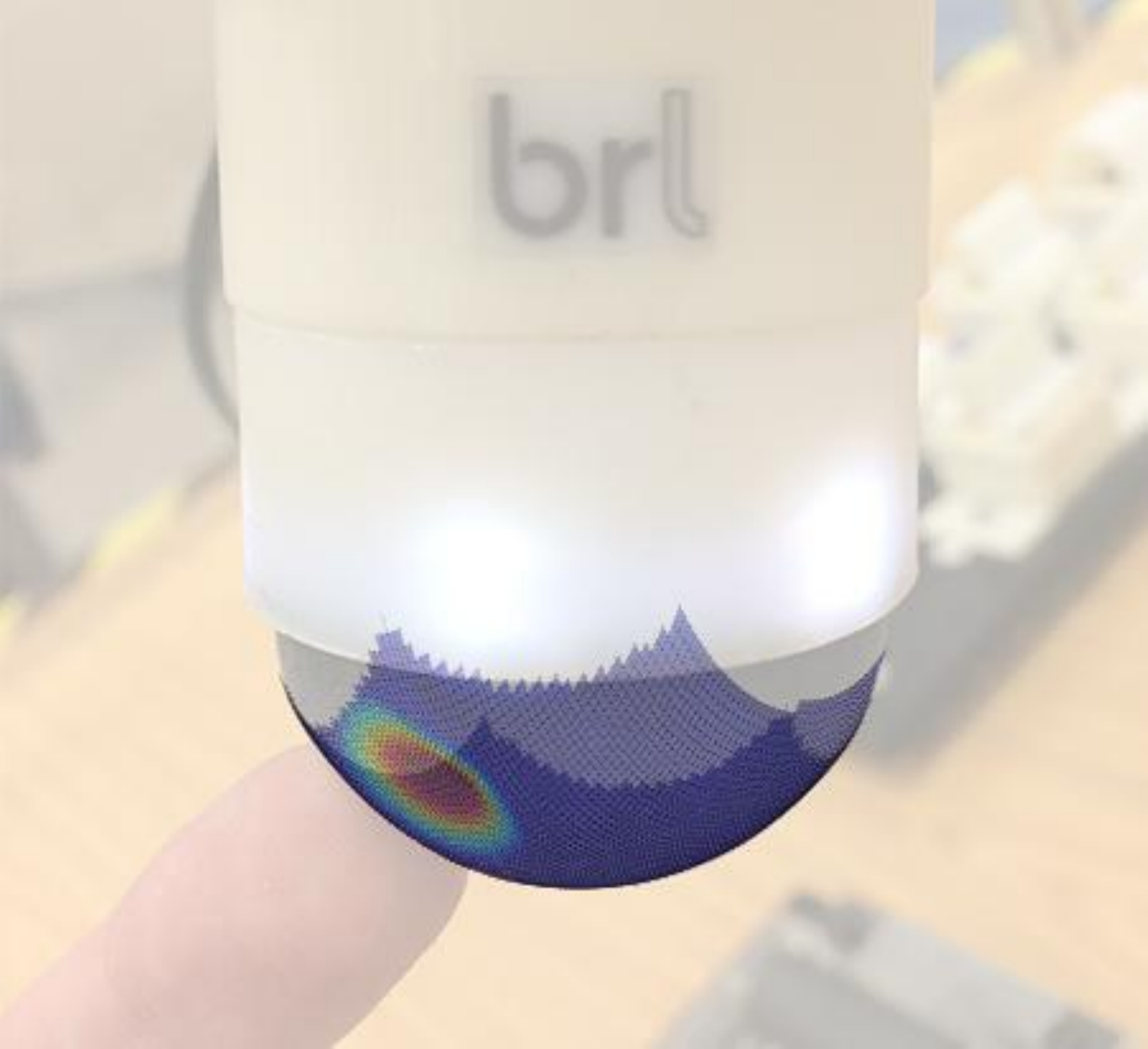}}\\
	\caption{The work presented here infers tactile features from TacTip sensory data by transducing a 3$^{rd}$ dimension of the data via Voronoi tessellation. This data can be used to produce a 3D reconstruction of the sensor tip. Shown here is sensory snapshot, reconstructed and shown with the sensor.}
	\label{fig:}
	\vspace{-1em}
\end{figure}  

This paper introduces a novel method for transducing deformation of the sensing surface by computing a third dimension for the data that is directly linked to the physical dimensions of touch, by utilising the natural region properties of Voronoi tessellation. This provides the ability to link the data to physical quantities such as displacement and contact location. We also demonstrate the inference of direction and magnitude of shear forces directly from the centroid locations from the sensor output. With these features it should be possible to utilise optical tactile sensors in classifier-free interpretation of contact that is both general and accurate.     

Voronoi tessellation is the mathematical principle of partitioning a plane into regions based on the distance between points and was defined and studied in a general $n$-dimensional case in 1908 by Georgy Voronoi. The Voronoi tessellation is widely used in science and technology; moreover, due to its versatility, the use of this method has been seen in art, science, geography, and many other fields. These diagrams are very effective for extracting extra information from point-based data sets and they also provide a visual adaptation to the data that enhances human visual interpretation of the information~\cite{Aurenhammer1991}.



\section{BACKGROUND AND RELATED WORK}

There are many sensors that are capable of detecting one or a combination of tactile modalities such as pressure, shear force, torque, slip, vibration, or temperature. Each tactile feature provides information about contact stimuli that can be used for control, perception, or exploration and with each sensor modality available the richness of data greatly aids in expanding the sensor functionality~\cite{Yousef2011,Kappassov2015}.



Surface displacement, pressure, and normal force are commonly used modalities in tactile sensors. With these, the necessary information is available for grasping an object with a sufficiently high force to maintain stability whilst restraining from applying forces that will either damage the stimulus, sensor, or manipulator. Sensors that can detect these features can use capacitive or resistive cells such as Harvards' TakkTile sensor~\cite{Tenzer2012}, CITECs' flexible fabric-based sensor~\cite{Buscher2015}, or force-torque sensors like the OptoForce.

Detection of shear forces is another tactile feature that many sensors exploit. The ability to detect shear forces are, along with normal forces, very useful for optimising the forces and quality of a grasp. Shear can also be used to identify friction coefficients and has been directly linked to slip detection (another important tactile feature)~\cite{Delhaye2014}. Tactile systems can be built around force torque sensors such as the OptoForce OMD can directly measure these features.

Contact location is useful from a control perspective, for example knowing the location of contact will enable tactile servoing to move to a different location or even to provide action on a stimulus for stabilising grasp or manipulation. Studies highlighting this important feature to extract from tactile sensors include work on the BioTac~\cite{Wettels2011} and an anthropomorphic tactile sensor~\cite{Koiva2013}.

An important aspect of the present work is the visualisation of tactile data. Other methods for visualising tactile data include snapshots of a tactile contact between the sensor and the stimulus which help to interpret the data~\cite{Bimbo2016,Goger2009}. Work by Cannata~\textit{et~al} has demonstrated a 2D somatosensory map to represent and visualise the contact on a tactile skin~\cite{Cannata2010}. 


The optical tactile sensor used in this paper is known as the TacTip. This sensor utilises an array of pins on the interior of a compliant skin to translate contact information into pin movement. The biomimetic aspect of the TacTip is that these pins perform a similar role to the intermediate ridges in the human fingertip. The end of intermediate ridges house the mechanoreceptors that traduce contact to nerve signals. These mechanoreceptor cells are known as the Merkal cells, whose output responds to stimuli similar to the displacement of the tracked pin heads of the TacTip~\cite{Cramphorn2017}.

From the initial development of the TacTip optical tactile sensor there have been a variety of methods implemented for interpretation the collected images. Initially, Chorley~\textit{et~al}, analysed the images using a velocity vector field and centroid tracking~\cite{chorley2009Development}. Most work with TacTip sensors has utilised pin tracking methods to record the pin positions from frame to frame. These pin positions are outputted as centroids of the white pin heads  in the form of $x$ and $y$ pixel coordinates. This data has been interpreted via Bayesian classification to achieve super-resolved acuity~\cite{lepora2015superresolution}, basic manipulation~\cite{cramphorn2016roller}, and contour following~\cite{Lepora2017}. Recent work on a cylindrical TacTip designed for detection of submucosal tumors during endoscopy utilised a vector representation of relative pin movement to colour a visualisation of the data; in addition a 3D reconstruction of the test environment was created from this method~\cite{Winstone2017}. This visualisation of the optical tactile sensor data helps the user to interpret the tactile data.    



\section{METHODS}

\begin{figure}[t!]
	\centering
	{\includegraphics[width=0.37\textwidth, trim = 0cm 0.9cm 0cm 0cm, clip]{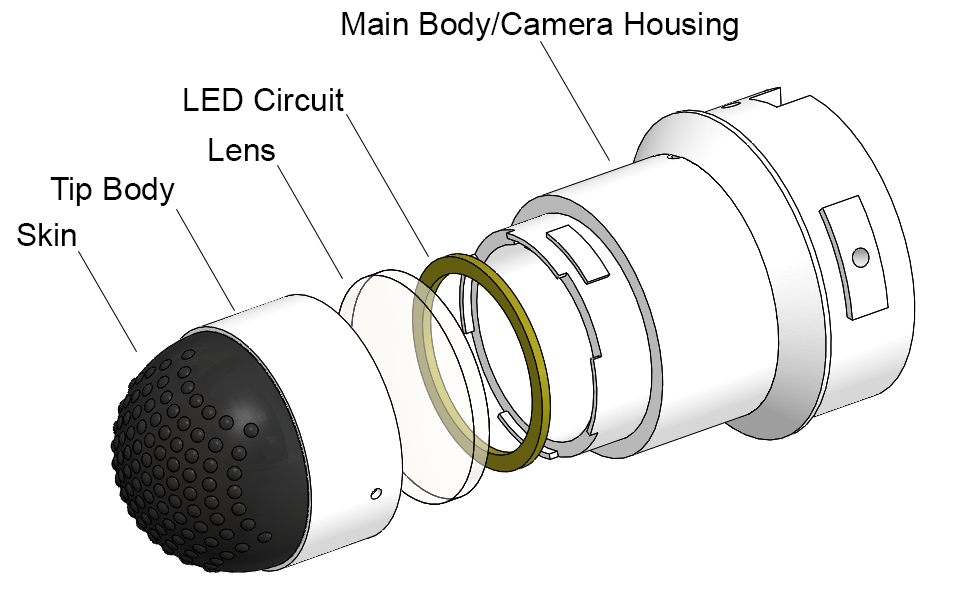}}\\
	\caption{Exploded view of a TacTip optical tactile sensor. The contact surface is referred to as the skin, this is a compliant 3D printed rubber that the interior of which is endowed with pins that are tracked by the camera}
	\label{fig:Tip}
	\vspace{-1em}
\end{figure}

\begin{figure}[t!]
	\centering
	{\includegraphics[width=0.35\textwidth, trim = 5cm 11.5cm 5cm 12cm, clip]{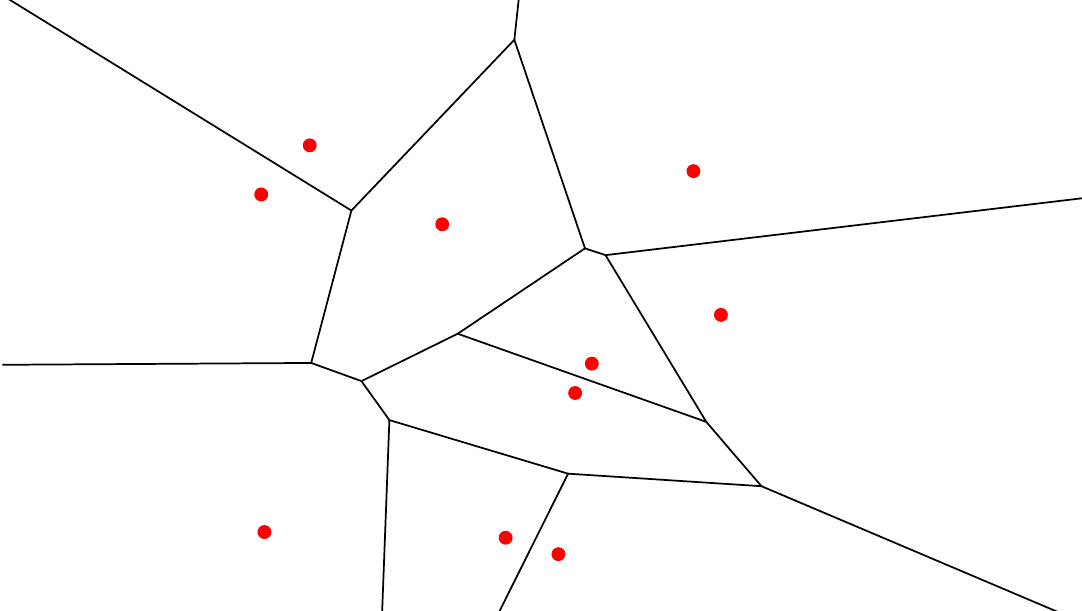}}\\
	\caption{Voronoi tessellation of 10 random points. Each point along an edge is equidistant from exactly two points, and each vertex is equidistant from at least three points. The Voronoi tessellation creates unbounded cells on the edge of the data, as lines extend to infinity with no other line to intercept.}
	\label{fig:Rand}
	\vspace{-1em}
\end{figure}

\begin{figure*}[t!]	
	\centering
	{\includegraphics[width=0.87\textwidth, trim = 0.4cm 0.1cm 0.5cm 0.7cm, clip]{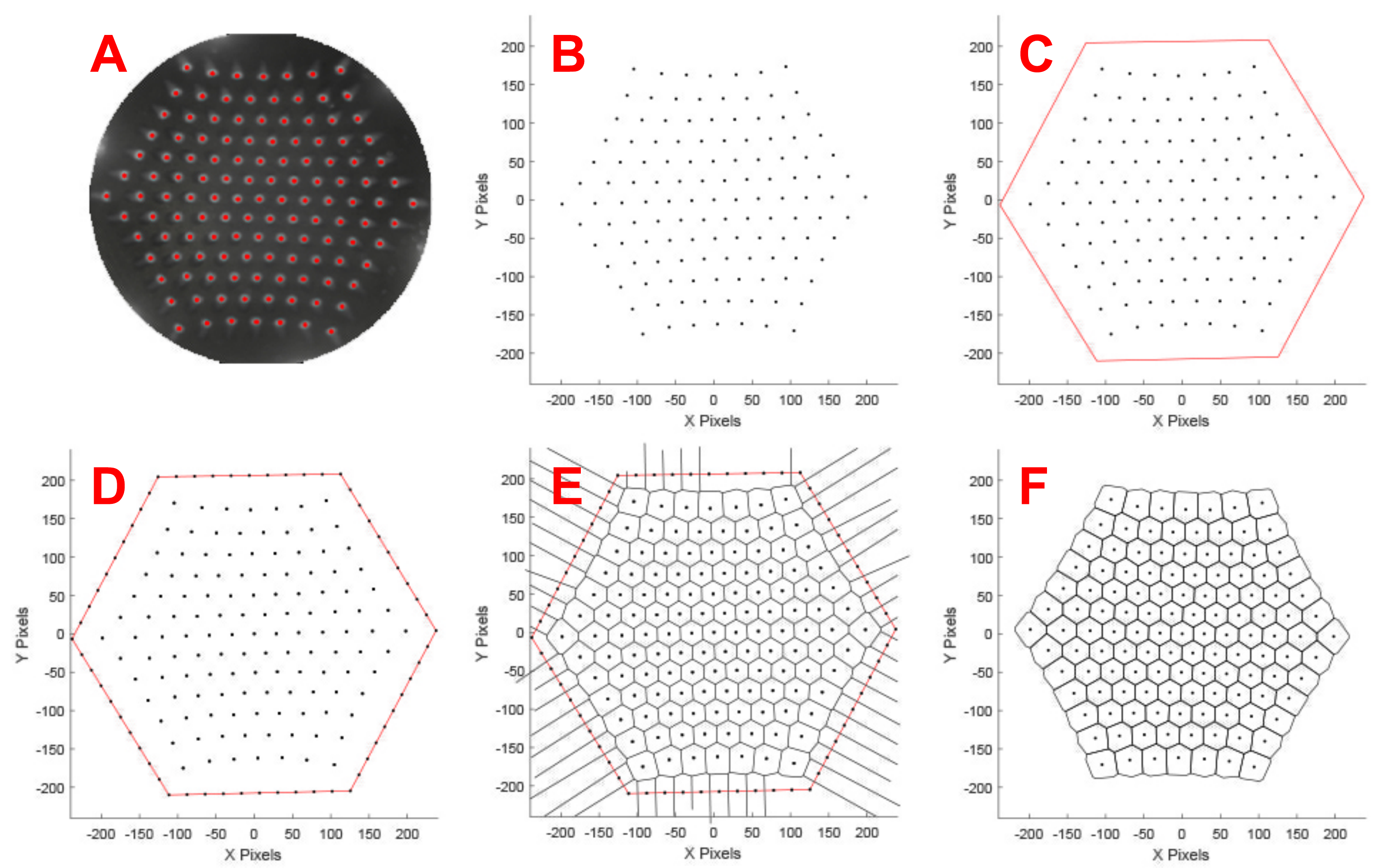}}\\
	\caption{To best use the Voronoi tessellation, a bound region is desired for each centroid collected from the sensor (A,B). As the outer centroids generate cells that are unbound a boundary is genarated to enclose all the centroids (C). Along this boundary artificial points are generated (D) allowing for bound cells on the real outer centroid (E). Finally all edges, longer than a threshold, and the artificial points are deleted from the data, resulting in the visual (F).}
	\label{fig:StoryBoard}
	\vspace{-1em}
\end{figure*}   

 
The TacTip optical tactile sensor is a biomimetic, low cost, 3D printed device which uses a compliant contact surface to extract information from stimuli~(Fig.\,\ref{fig:Tip}). This surface, or skin, is a rubber-like material printed by a multi-material 3D printer (Objet) and is filled with a optically clear gel to support the structure and provide compliant properties. The interior of the skin has an array of pins with a white plastic cap. Tracking of these pins is performed by a plug and play web cam (Microsoft life-cam) that is located above the contact part of the sensor that will be referred to as the tip. The arrangement of pins optimises pin density and tracking. The layout is a 2D hexagonal projection from the perspective of the camera. The projection spaces pins at $\sim$3\,mm, with each pin having a length of $\sim$2\,mm. This means that the pins are not evenly spaced on the sensor surface but instead generate a regular and evenly spaced pattern for the camera, which supports robust pin tracking~\cite{Cramphorn2017}.

\subsection{Voronoi Tessellation}

Voronoi tessellation is the mathematical principle of partitioning a plane into regions based on the distance between points on that plane. The regions created by this tessellation are directly associated with the point each is closest to. Rules in Voronoi tessellation are that each point along an edge is equidistant from exactly two points, and each vertex is equidistant from at least three points. Edge lines will tend to infinity if there is no other edge to intercept and create a vertex with. These infinite edges tend to occur at the edge of the point data and will results in unbounded regions.     

\subsubsection{Construction of Voronoi Cell Data}

As mentioned, Voronoi tessellation is a powerful tool for extracting information from a plane defined by a series of points. Due to its usefulness there are predefined computational tools for creating this tessellation on a set of input points.

To make the best use of the Voronoi tessellation, it is desirable to have bound cells for every centroid recorded by the sensor (Fig.\,\ref{fig:StoryBoard}A,B). The outer centroids generate cells that are unbound, to solve this a boundary (Fig.\,\ref{fig:StoryBoard}C) fitted with artificial centroid points (Fig.\,\ref{fig:StoryBoard}D enclosing the real data. By doing this the unbound Voronoi cells are around artificial points rather than real centroid data (Fig.\,\ref{fig:StoryBoard}E). The point along the boundary have a density higher than that of the inner points allowing for a more stable outer cell shape when the Voronoi is applied. Removing edges longer than those in the centre and deleting artificial points from the data leaves Voronoi cell structure for all recorded centroids (Fig.\,\ref{fig:StoryBoard}F).
 
The area of the bound cells produced by the Voronoi tessellation provide new information that is valuable for tactile perception. Specifically the change in each regions areas relates to sensor compression when they increase and swelling (due to stable internal volume) when they decrease. 

\subsubsection{Inference of key Tactile Features}
\label{InfTac}

Each of the tactile features - pressure, shear, and contact location - provides useful information about contacted stimuli. The combination of these sensory modalities can be used to provide a rich information base for perception and manipulation of objects.         

\noindent\textit{a)~Surface Displacement, Pressure, and Normal Force:}~The normal force exerted on the tip will be proportional to the extent of deformation the sensor undergoes. Thus being able to infer the level of deformation is important. The area values for each Voronoi cell can be used as a $z$-dimension in a surface fit via cubic interpolation. This 3-dimensional surface produced a proportional approximation of surface deformation and indention. The integral of this surface is the volume of deformation, a value that can be easily calibrated to mechanical units of force or displacement and combined with the area of contact for pressure.       

\begin{figure}[t!]
	\vspace{-1em}
	\centering
	{\includegraphics[width=0.45\textwidth, trim = 0cm 0cm 0cm 0cm, clip]{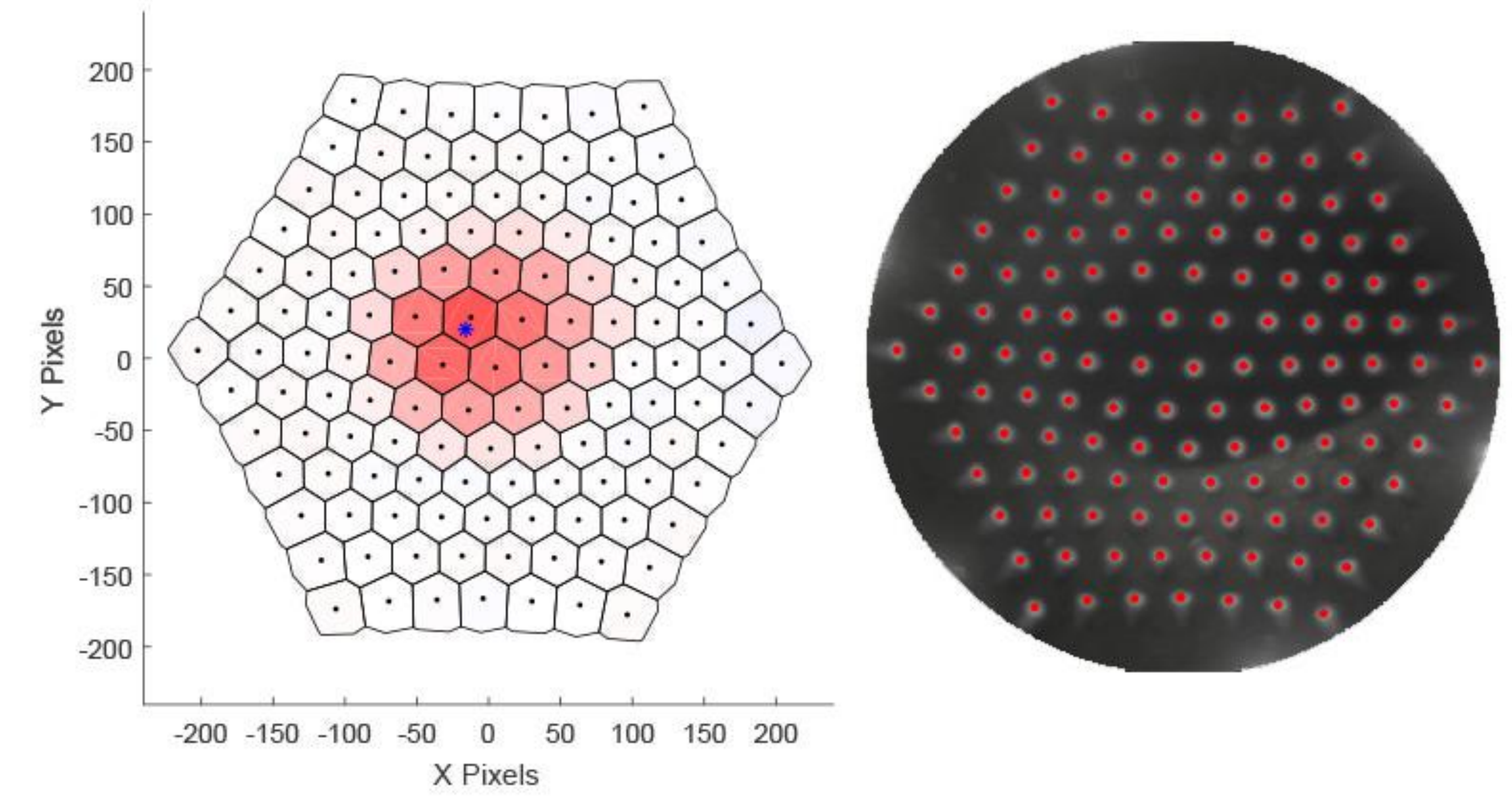}}\\
	\caption{Visualisation (left) where Voronoi cells are coloured red proportional to the increase in cell area. This helps to visually interpret TacTip data. The blue asterix is the estimated centre of contact at the maximum of a smooth surface fit  (Fig.~\ref{fig:3D}). The frame prior to visualising (right), emphasises the benefits of the Voronoi cell areas for visually interpreting the data.}
	\label{fig:Normal}
	\vspace{1em}
	\centering
	{\includegraphics[width=0.4\textwidth, trim = 3.4cm 10cm 4cm 13.3cm, clip]{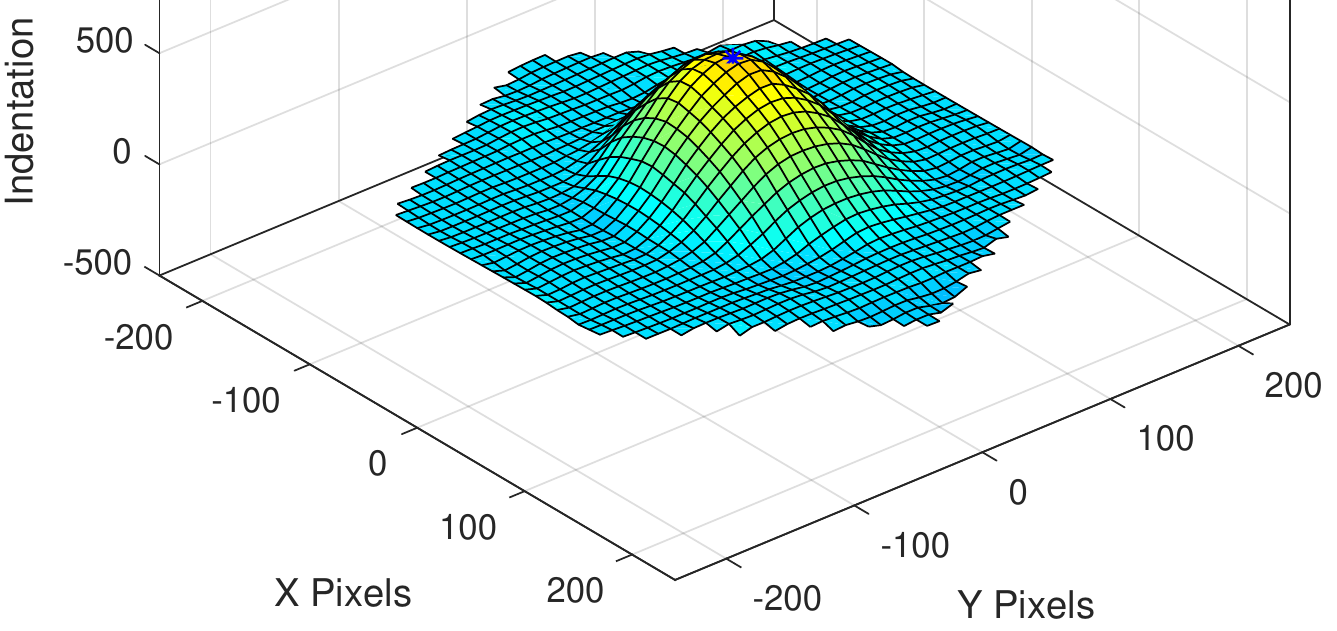}}\\
	\caption{Using centroid $x$ and $y$ values along with the Voronoi cell areas as the $z$ values, a 3D surface fit can be applied to the data via cubic interpolation. This fit allows for a proportional reconstruction of sensors deformation in both compression and expansion. The maximum $z$ value along this reconstruction is the local point of greatest deformation, and in simple contacts, such as a finger press or probe, is the centre of contact.}
	\label{fig:3D}
	\vspace{-1em}
\end{figure}

\noindent\textit{b)~Contact Location and Multi-Contact Detection:}~The 3D surface data created to infer deformation can also be utilised to determine an approximation of the contact locations. Contact location provides information that can be used to asses many aspects of how the sensing surface is interacting with an object. Inferring contact locations can be done by finding the maxima on the surface where the contact location is assumed to be the regions of greatest deformation.

This extend to multiple contacts when the local maxima are taken into account. To find local maxima on the surface, the data is treated as an image ($n\,\times\,m$ matrix of values). The regional maxima of this matrix are connected components of pixels (array values) with a constant intensity value with external boundary pixels all with values less than these. The result is a binary $n\,\times\,m$ array where  1 identifies the regional maxima and all other array values are 0 (implemented by the  \textit{imregionalmax} function in MATLAB). This returns the coordinates of local maxima in the data, of which all maxima below a contact threshold are removed.  


\noindent\textit{c)~Local and Global Shear:}~Inferring shear from the TacTip is achievable by looking at the velocity vectors of each centroid from frame 1 (calibration frame) to the current frame, as noted in the original TacTip paper~\cite{chorley2009Development}. Each centroid has an associated vector that measures local shear. These local shears can be grouped to average regions of shears or they can be combined globally to compute the overall global shear on the tip. Naturally this vector can be broken down into an estimation of shear direction and shear magnitude. 

   
\subsection{Experimental Procedure} 
\label{Exp}


Validating the tactile modalities inferred here requires a system that can control the sensor position with high precession. Following prior work using the TacTip, the sensor is mounted on a 6 dof robot arm (IRB 120, ABB Robotics) with absolute position repeatability 0.01\,mm \cite{Cramphorn2017,lepora2015superresolution,Ward-Cherrier2016}.

The raw values produced through the shear magnitude and surface displacement modalities, mentioned in the above methods, are in the units of the methods used to infer them (e.g. mm). Calibrating the values to mechanical units allows for better understanding of the system and allows comparison with other sensors. Calibration data is collected on a flat level surface using the robot arm. For indentation depth, recordings of 10 sample frames per step, at depths of 0\,mm to 5\,mm in steps of 0.1\,mm. For the physical arm set-up, used here, it is best to calibrate against the displacement of the tip; which will be non-linearly related to the pressure (N\,m$^{-2}$). Calibrating shear magnitude is done with 10 sample frames per step, collected for shear distances of 0\,mm to 2\,mm in steps of 0.1\,mm. The resulting data (recordings against telemetry) of values is fitted via linear interpolation.

The validation of shear direction is performed by pressing the sensor into a flat level object approximately 3\,mm and then creating a lateral move of another 2\,mm. This shears the tip without generating slip movements, that would distort the readings for this validation. This is repeated 36 times in 10$^{\circ}$ increments from 0$^{\circ}$ to 350$^{\circ}$. The sensor output and known shear directions are then compared.        
  

\section{Results}
%
%
%


\subsection{Surface Displacement, Pressure, and Normal Force}

It can be clearly seen that when contact is made with the sensor surface that the Voronoi cells increase in area (Figure~\ref{fig:Normal}; showing centroids and camera image with visualisation). For visualisation purposes this is enhanced by assigning a colour shading to the cells based on the percentage change of the area. Shades of red are associated with sensor compression (increase in Voronoi cell area) and shades of blue for sensor expansion (decrease in Voronoi cell area).

Plotting the surface fit, (Section~\ref{InfTac}) for inferring the contact location, and estimating a proportional surface displacement, produces the 3D figure shown in Fig.\,\ref{fig:3D}. The maximum value on this fit is marked with a blue star (shown on Fig.\,\ref{fig:Normal} for comparison). Hence the contact location can be inferred in the format of $(x,y,z)$, where $z$ is a value proportional to the maximum surface indentation. The volume of this surface is calculated by integration, and is proportional to an estimate of surface displacement, which in turn can be associated with the applied force on the surface and thus with the pressure exerted. The height of indentation is an amplification of real deformation, and could be calibrated to produce mechanical values for displacement, force, or pressure. Figure~\ref{fig:Normal} also shows a ring of slightly concave surface around the peak, which represents an outer region of slight compression around the contact.

 
\subsection{Contact Location and Multi-Contact Detection}

\begin{figure}[t!]
	\centering
	{\includegraphics[width=0.45\textwidth, trim = 0cm 0cm 0cm 0cm, clip]{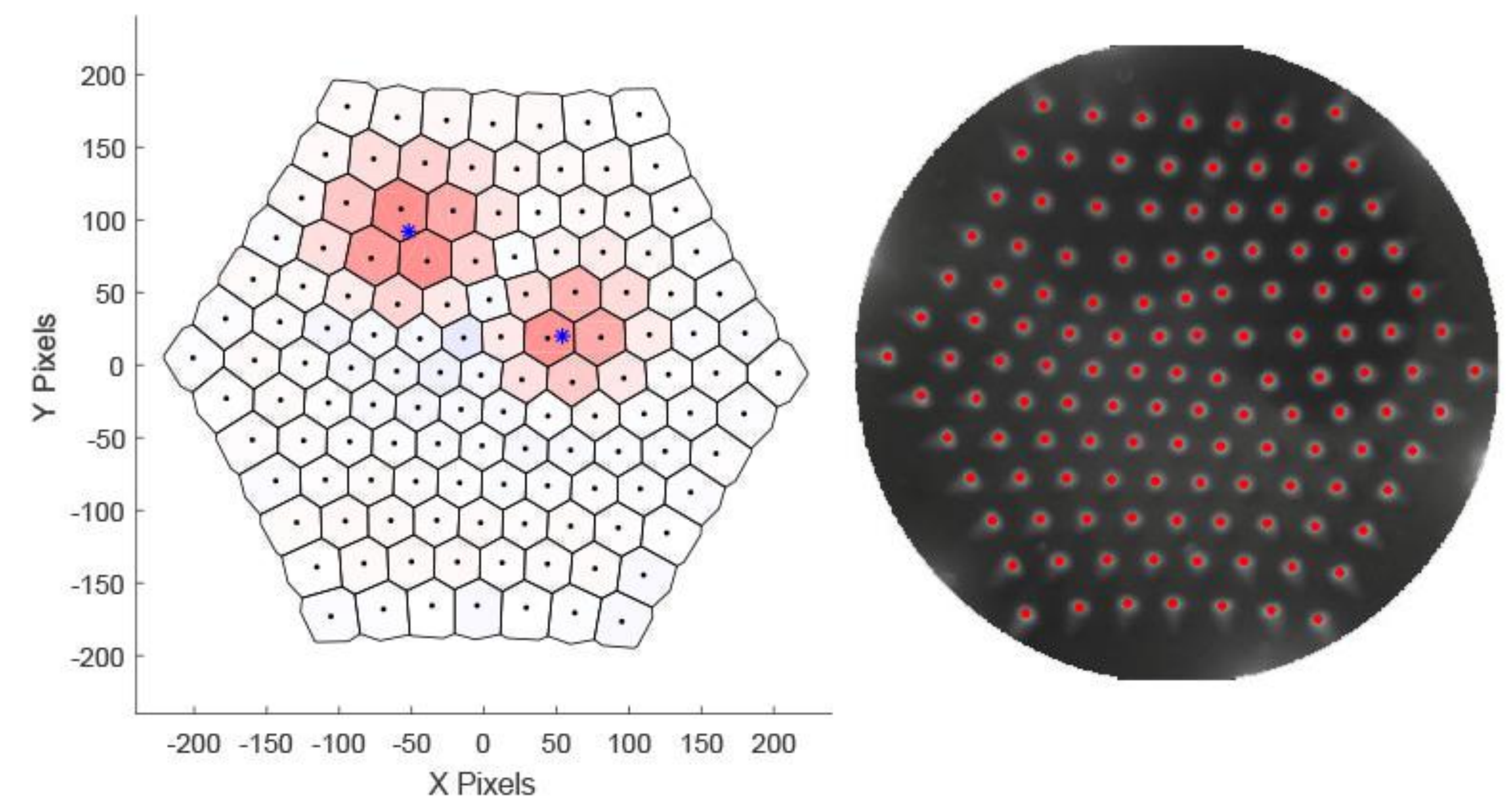}}\\
	\caption{Further interpretation of the fit function can be done to detect local maxima. This means that the multiple, distinct, contact points can be identified on the sensor. Here the visual (left) shows the two contact sites with blue star marking the estimated centre of contacts. The centroids and camera image for this frame are also shown (right).}
	\label{fig:Multi}
	\vspace{-1em}
\end{figure}  
\begin{figure}[t!]
	\centering
	{\includegraphics[width=0.45\textwidth, trim = 0cm 0cm 0cm 0cm, clip]{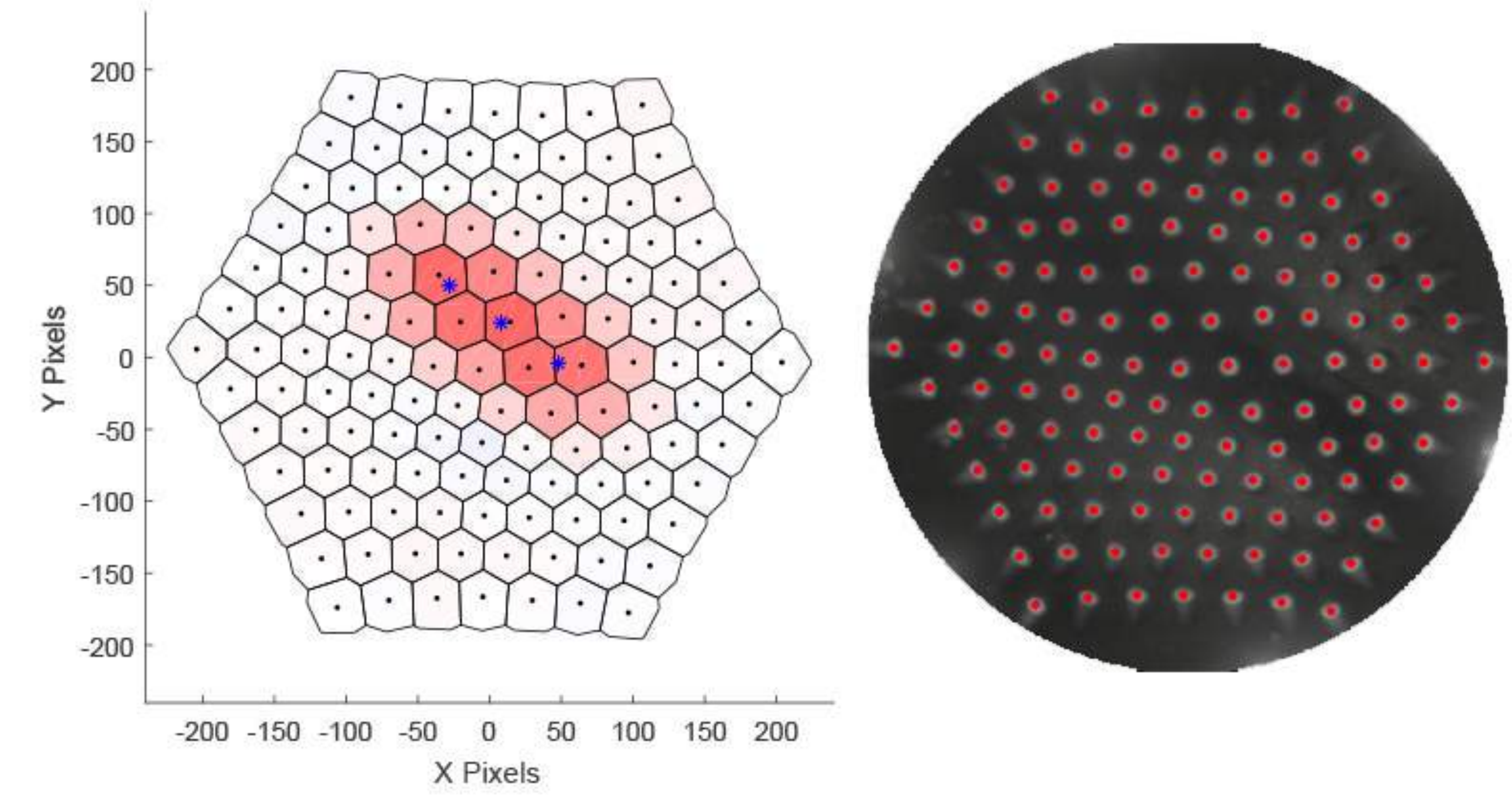}}\\
	\caption{An example of an elongated contact on the sensor. The identified contact point number greater than the number of stimuli, although this may be useful for inferring stimulus orientation.}
	\label{fig:Multilong}
	\vspace{-1em}
\end{figure}  

Being able detect multiple regions of contact can be used to perceive multiple stimuli or a stimulus with an interesting shape. The maxima approach previously discussed and demonstrated in the pressure section of the results is easily extendible to multiple contacts. This is shown in Fig.\,\ref{fig:Multi} where two distinct points of contact are inferred and then marked (as blue stars) on the visualisation.

The number of distinct contact locations that can be identified is limited by the nature of the contact. For example point contacts from simple probes that are sufficiently spaced ($\sim$4\,mm) can easily be identified as separate objects. Although more complex probes, such as flat or long stimuli, create a elongated region of indentation where multi maxima are found, even though there is only a single contacting stimulus. In addition to this, flat objects create a ring of contact point around the centre of contact, due to the surface fit being a flat plane rather than a hemisphere like the sensor. Interestingly the long thin object generate contact locations that follow the length of the contact, which may be used for the identification of object/edge orientation by taking the direction of a line that connects these points (Fig.\,\ref{fig:Multilong}).  

\subsection{Local and Global Shear} \label{LShear}

\begin{figure}[t!]
	\centering
	{\includegraphics[width=0.45\textwidth, trim = 0cm 0cm 0cm 0cm, clip]{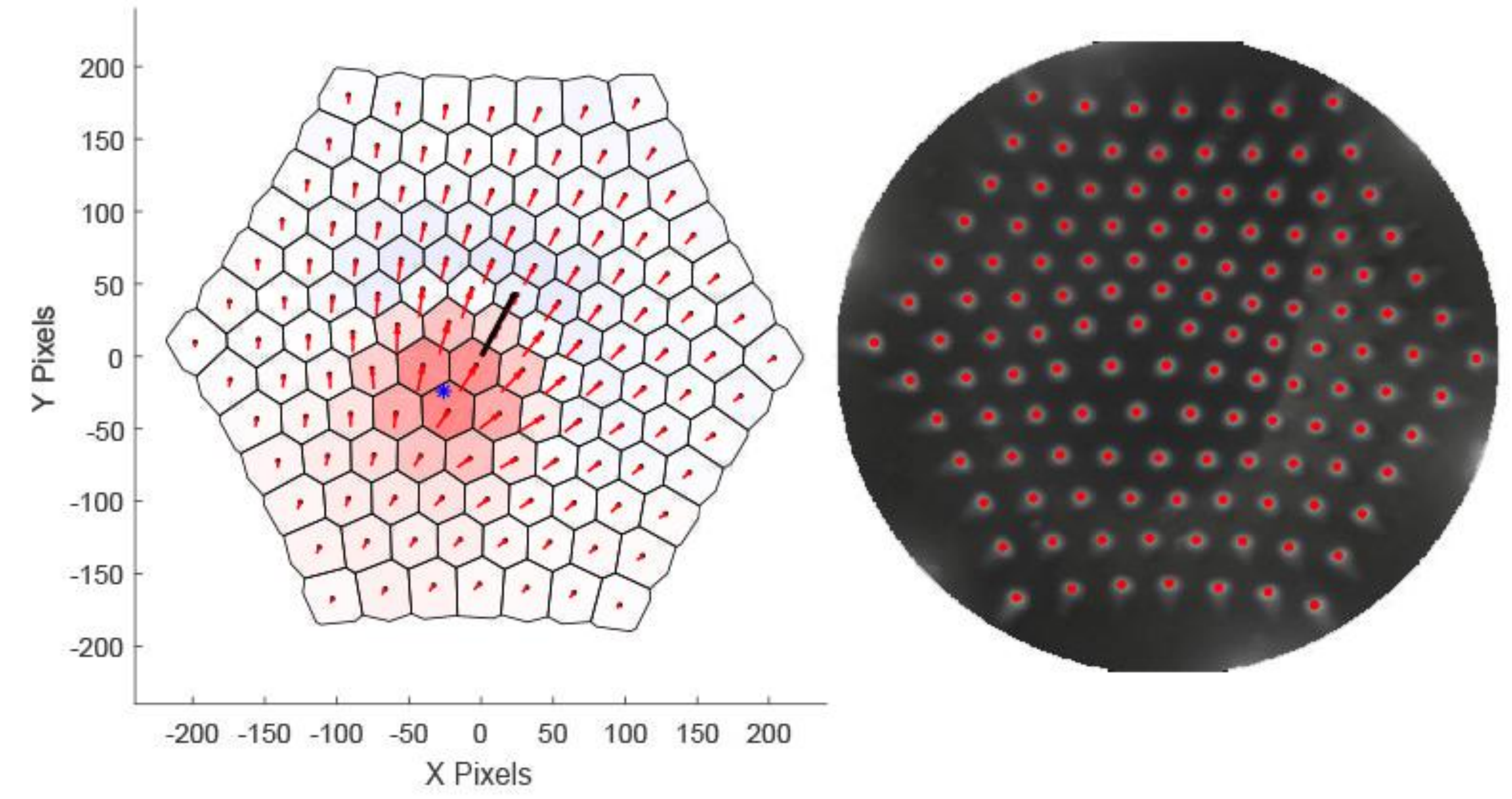}}\\
	\caption{The direction and magnitude of the centroid motion vectors are associated with the shear strain on the sensor. Assuming that the velocity vectors for each of the centroids is a local shear (marked as red vectors on the visual (left)), then the average magnitude and direction of all centroid vectors will produce an average global shear vector (black arrow). Again the frame from the camera and the centroids are shown for reference (right).}
	\label{fig:shear}
	\vspace{-1em}
\end{figure}

Under shear forces the centroids of the pin move relative to the shear direction, whereas under normal forces the centroids move radially from the point of contact. The global value of shear is the mean of all local shears (individual centroid motion), which will result in a non zero magnitude and direction when the contact applies shear forces. These local and global shear vectors are plotted on the visualisation in Fig.\,\ref{fig:shear}. Here it can be seen that the normal component of contact is still detectable, allowing for both normal and shear forces to be detected simultaneously. The Voronoi method is not used to calculate the shear, due to the fact that it is directly inferred from the centroid data; however, there is a clear increase in sensor expansion (blue shaded cells) in a crescent around the contact site and in the direction of shear.     

\begin{figure}[b!]
	\centering
	{\includegraphics[width=0.4\textwidth, trim = 4cm 12cm 4cm 12.63cm, clip]{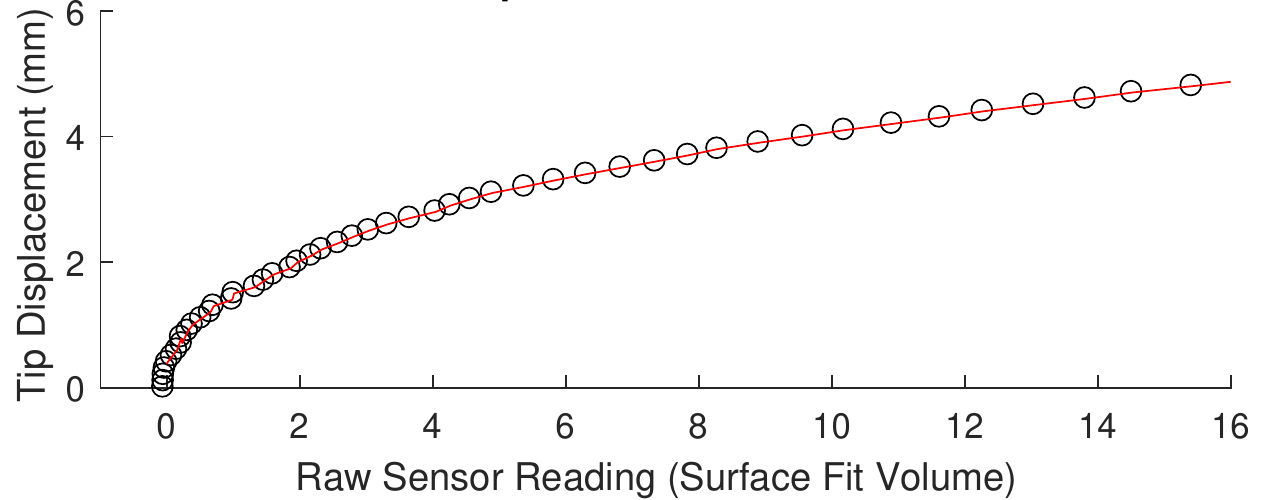}}
	\vspace{0em}
	\centering
	{\includegraphics[width=0.4\textwidth, trim = 4cm 12cm 4cm 12.63cm, clip]{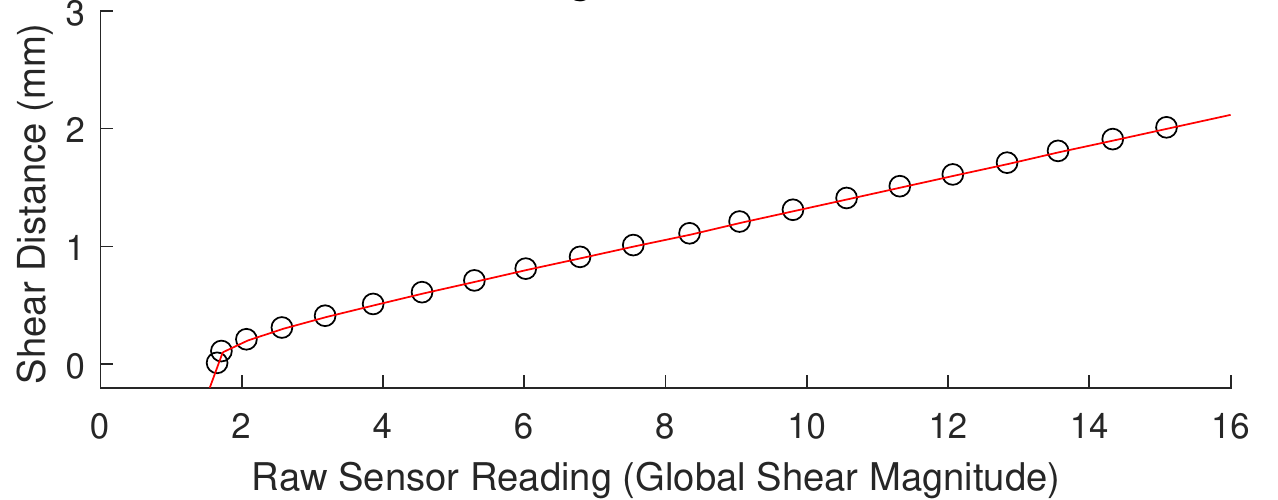}}\\
	\caption{Calibration of the sensor is achieved by producing a fit between the output values of the inferencer and the telemetry values of a robot arm (see Sec.\,\ref{Exp} for detail). The resulting calibration fits can be seen above for displacement (top) and for the shear magnitude (bottom)}
	\label{fig:Calibration_Curves}
	\vspace{0em}
\end{figure} 

During the validation for inferring shear direction, the shear magnitude and tip displacement are also inferred. This shows that the modalities can be simultaneously extracted from the sensor data. Applying the calibration, described in Section~\ref{Exp}, to the sensor output values produces a calibration curve for both shear magnitude and tip displacement. These represent the relationship between the sensor output and mechanical units of the arm telemetry (Fig\,\ref{fig:Calibration_Curves}). 
	
The calibration curve for tip displacement (Fig.~10, top) shows a non-linear relationship between displacement and sensor output. This suggests that the sensor is less sensitive to low tip displacement than it is to higher displacements. 

In contrast, the fit for shear magnitude is linear (Fig.~10, bottom), as the centroid positions when calibrated form a vector linearly related to the shear distance. The exception to this linearity is the first recording in the calibration set which has a higher inferred value than the other data would suggest. As this value has zero shear magnitude, it could be assumed that the sensor recording for no movement was in fact under a preloaded shear, possibly caused by divergence in angle between the normal of the calibration surface and the approach of the robot arm.      

\begin{figure}[t!]
	\centering
	{\includegraphics[width=0.33\textwidth, trim = 5cm 9cm 5cm 9cm, clip]{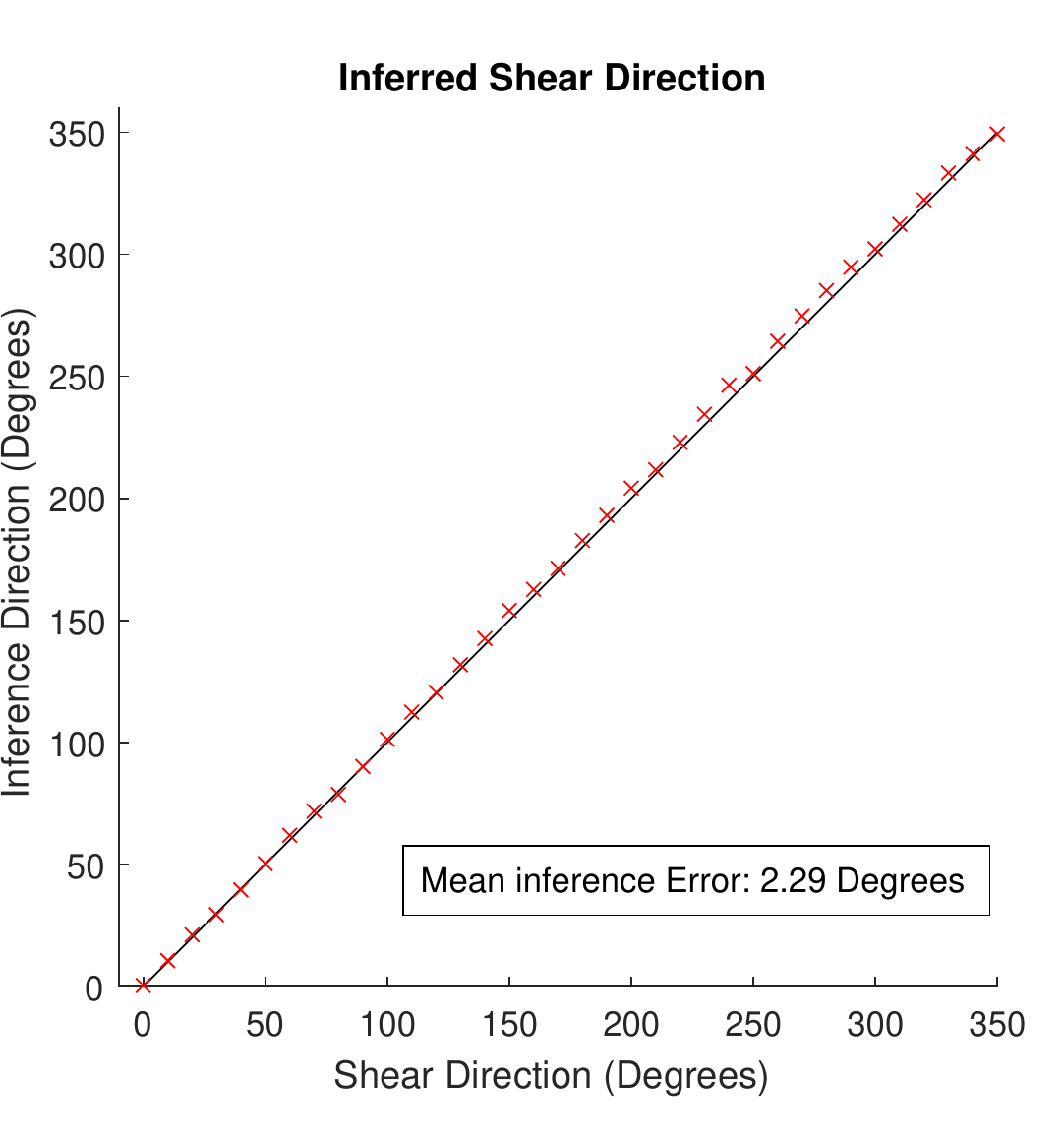}}\\
	\caption{Experimental validation of inferred shear direction is done by recording robot arm telemetry. The test is carried out for shears in 10$^{\circ}$ increments for a full circle. The resulting inferred values (plotted here) shows that the inferred direction is accurate to within an average of $\sim2.3^{\circ}$. Thus, without training, the direction of shear can be accurately inferred.}
	\label{fig:Error}
	\vspace{-1em}
\end{figure}

Validating shear direction is done by comparing inferred angles with telemetry from the arm used for controlling the sensor position (Fig.~11). Shears are tested at 36 different direction in 10$^{\circ}$ increments around a central location. The error in inference of direction is shown in Fig.~\ref{fig:Error} showing that the system can infer shear direction to an average of ${\sim~\!2.3^{\circ}}$. This validation is shown in a video associated with this paper, in which the Voronoi visualisation and contact location are seen along with tip displacement, shear magnitude, and the real and inferred shear directions.    


%
%
%
%

\section{Discussion}

This paper demonstrates a novel method for inferring tactile features from optical tactile sensor data, for generalising the use of these sensors, by producing useful information without the need for black-box classification or regression. 

We have found that the natural features produced by performing Voronoi tessellation on the sensor data creates a tool for generating a third dimension that can be used to transduce the deformation of the surface. This transduced deformation can be utilised to produce a data visualisations that provides information useful for the inference of important tactile features. For example, deformation and contact location can be directly inferred, which would be useful for control tasks like tactile servoing~\cite{Cannata2010}. We also show that it is possible to infer shear direction and magnitude directly from the centroid data due to the inherent translational information encoded in the changes of centroid positions. 


The inference of shear direction is validated against arm telemetry and showed predications to an average of $\sim2.3^{\circ}$. Although it is likely that the error in shear direction would be lower for classifier methods previously used, it is clear that as there is no training required and the shear can be interpreted under different pressures, the generality of the method will be of benefit for future work.      

The methods proposed here are the initial attempts to infer tactile features from the TacTip tactile data. There is the potential to expand the number of features inferable from the data to include inference of torque, slip, and the shape and orientation of contact stimuli. Having data from all of these modalities will further enhance the usability and versatility of the sensor. One example for this would be to use the inference of these features to control a robot hand, a task that is challenging to do with classifiers due to the lower accuracy of hands in comparison to robot arms~\cite{Goger2009}. 



The visualisation of the tactile data, via the Voronoi tessellation, also proved to be a valuable output of this work, as the new visualisations make it much easier to interpret the tactile data by eye. The quote from a leading survey on Voronoi diagrams highlights this as an underlying property of the tessellation that is clearly demonstrated in these visual representations: \textit{`Human intuition is often guided by visual perception. If one sees an underlying structure, the whole situation may be understood at a higher level.'}~\cite{Aurenhammer1991}.

\end{document}